\documentclass[letterpaper, 10 pt, conference]{ieeeconf}  % Comment this line out
\IEEEoverridecommandlockouts                              % This command is only
\usepackage[utf8]{inputenc}
\usepackage[T1]{fontenc}
\usepackage{graphicx}
\usepackage{booktabs}
\usepackage{multirow}
\title{\LARGE \bf
From Theory to Application: Fine-Tuning Large EEG Model with Real-World Stress Data 
}

\author{Siwen Wang$^{1}$, Shitou Zhang$^{2}$, Wan-Lin Chen$^{3}$, Dung Truong$^{4}$ and Tzyy-Ping Jung$^{5}$% <-this % stops a space
\thanks{*This work was not supported by any organization}% <-this % stops a space
\thanks{$^{1}$S. Wang is with the Department of Mathematics and Computer Science, Qingdao Academy, Qingdao, Shandong, China
        {\tt\small wangsiwen at qdzx.net}}%
\thanks{$^{2}$S. Zhang is with the Department of Information Engineering, the Chinese University of Hong Kong,
        {\tt\small stzhang at link.cuhk.edu.hk}}%
\thanks{$^{3}$W. Chen is with the Department of Bioengineering, University of California, San Deigo,
        La Jolla, CA 92037, USA
        {\tt\small wac006 at ucsd.edu}}%
\thanks{$^{4}$D. Truong is with the Swartz Center for Computational Neuroscience, University of California, San Diego,
        La Jolla, CA 92037, USA
        {\tt\small dutruong at ucsd.edu}}%
\thanks{$^{5}$T. Jung is with the Swartz Center for Computational Neuroscience, University of California, San Diego,
        La Jolla, CA 92037, USA
        {\tt\small tpjung at ucsd.edu}}%
}

\begin{document}

\maketitle
\thispagestyle{empty}
\pagestyle{empty}

%%%%%%%%%%%%%%%%%%%%%%%%%%%%%%%%%%%%%%%%%%%%%%%%%%%%%%%%%%%%%%%%%%%%%%%%%%%%%%%%
\begin{abstract}

Recent advancements in Large Language Models have inspired the development of foundation models across various domains. In this study, we evaluate the efficacy of Large EEG Models (LEMs) by fine‑tuning LaBraM, a state‑of‑the‑art foundation EEG model, on a real‑world stress classification dataset collected in a graduate classroom. Unlike previous studies that primarily evaluate LEMs using data from controlled clinical settings, our work assesses their applicability to real‑world environments. We train a binary classifier that distinguishes between normal and elevated stress states using resting‑state EEG data recorded from 18 graduate students during a class session. The best‑performing fine‑tuned model achieves a balanced accuracy of 90.47\% with a 5‑second window, significantly outperforming traditional stress classifiers in both accuracy and inference efficiency. We further evaluate the robustness of the fine‑tuned LEM under random data shuffling and reduced channel counts. These results demonstrate the capability of LEMs to effectively process real‑world EEG data and highlight their potential to revolutionize brain-computer interface applications by shifting the focus from model‑centric to data‑centric design.

\end{abstract}

%%%%%%%%%%%%%%%%%%%%%%%%%%%%%%%%%%%%%%%%%%%%%%%%%%%%%%%%%%%%%%%%%%%%%%%%%%%%%%%%
\section{INTRODUCTION}

Recent success in Large Language Models (LLMs) has spurred the creation of foundation models across different domains \cite{moor2023foundation}\cite{yuan2021florence}\cite{yu2022coca}. A foundation model is pretrained on broad data and can be adapted (e.g., fine-tuned) to a wide range of downstream tasks \cite{bommasani2021opportunities}. Large EEG Models (LEMs), a type of foundation model trained on large-scale EEG data, have shown success in various downstream tasks, including medical diagnosis \cite{yuan2024brant}, affective computing \cite{jiang2024large}, and brain-computer interface (BCI) applications \cite{cui2023neuro}. 

Traditional BCI models struggle with generalization, requiring significant time and effort to design unique model architectures for each application. Recent advances in foundation EEG models provide a data-centric alternative for developing BCI applications, allowing researchers to focus on curating high-quality data rather than manually designing model architectures. This paradigm has already demonstrated strong performance in downstream tasks, sometimes outperforming expert-designed models \cite{jiang2024large}\cite{chen2024eegformer}\cite{kim2024eeg}. 

However, most LEMs have only been tested on datasets collected in controlled settings, such as hospitals and research labs, leaving their applicability to real-world data uncertain. In this study, we fine-tuned LaBraM \cite{jiang2024large}, one of the state-of-the-art LEMs, on our stress dataset collected in a real-world classroom. Our results show that the model achieved a balanced accuracy of 90.47\%  on a binary stress classification task using just a 5-second window. The fine-tuned LEM significantly outperformed traditional stress classifiers in classification accuracy and inference efficiency, confirming its ability to interpret real-world EEG data. Additionally, our results in evaluating the robustness of the fine-tuned LEM underscore the current model's limitations and highlight directions for future optimization.

\section{Method}

\subsection{Large EEG Model}

We employ LaBraM, a state-of-the-art LEM, as the foundation model to explore its potential for efficient downstream generalization to real-world stress data. LaBraM was pretrained on a mixed collection of EEG recordings (totaling over 2,500 hours) and follows a two-stage training pipeline: (1) training a neural tokenizer to convert EEG signals into discrete neural tokens by reconstructing their Fourier spectra, and (2) training LaBraM with masked token reconstruction to learn transferable features. Although LaBraM has shown strong performance on standard benchmarks, including TUAB and TUEV \cite{obeid2016temple}, it has yet to be evaluated on stress data. Thus, we choose LaBraM as the base model to investigate its downstream adaptability to stress detection.

\subsection{Downstream Fine-tuning Dataset}

\subsubsection{Participants}
Eighteen graduate students ($24 \pm 1.2$ years old; 10 male, 8 female) from National Chiao Tung University, Taiwan, participated in this study during the first semester of the 2014–2015 academic year.

\subsubsection{Experimental Design}
We collected data in a real-world classroom while the class was in session. EEG data were recorded using a Compumedics Neuroscan system at a sampling rate of 1000 Hz. We placed thirty sensors on the participants' scalps using a 32-channel Quick-Cap. Before each data collection session, participants completed the DASS survey to assess levels of stress, anxiety, and depression. We then collected a 5-minute eyes-open resting-state EEG as the baseline. A total of 92 recordings (21 labeled as elevated stress and 71 as normal stress) from these 5-minute sessions were used in subsequent analyses.

\subsection{Pre-processing}

We perform a sequence of signal processing techniques to ensure data quality. First, we apply a band-pass filter between 1 and 50 Hz to reduce high-frequency artifacts and some eye-related activity. Next, we apply Artifact Subspace Reconstruction (ASR) \cite{chang2018evaluation} to eliminate high-amplitude artifacts. Finally, we use Independent Component Analysis (ICA) \cite{makeig1995independent} to separate statistically independent components from the EEG data and automatically reject noncortical independent components (ICs) using EEGLAB's automatic IC labeling tool. We set the rejection threshold to 80\%; that is, the algorithm rejects an IC automatically if the probability of being a brain component falls below 80\%.

\subsection{Further Data Processing Before Fine-tuning}

\begin{figure}[htbp]
    \centering
    \includegraphics[width=0.48\textwidth]{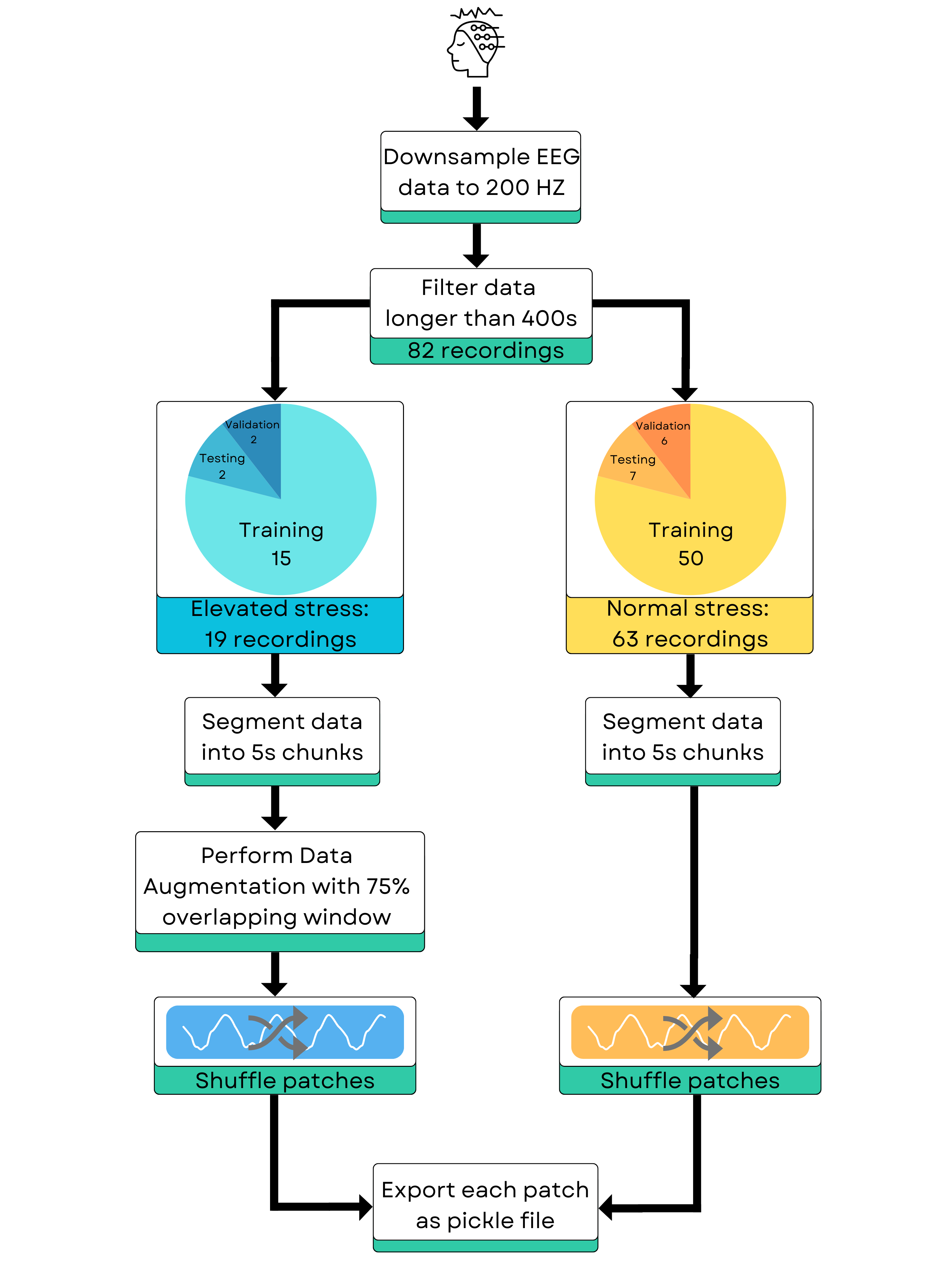}
    \caption{Further Data Processing Pipeline}
    \label{fig:further data process}
\end{figure}

We conduct additional data processing before fine-tuning (Fig.~\ref{fig:further data process}). First, we downsample all EEG data to 200~Hz to align with the LaBraM model configuration. Then, we discard any recordings longer than 400 seconds to filter based on recording length. Recordings that exceed this duration typically result from poor signal quality during data collection, often caused by the assistant moving around the EEG cap to reduce impedance. This additional noise likely degrades model training and prevents proper convergence. 

After filtering, 82 recordings remain: 19 in the elevated stress group and 63 in the normal stress group. Next, we divide the elevated stress and normal stress groups into training, validation, and testing sets in approximately 80\%, 10\%, and 10\% proportions, respectively. For the elevated stress group, we assign 15 recordings to training and 2 each to validation and testing. For the normal stress group, we use 50 recordings for training, 6 for testing, and 7 for validation.

We then segment each subject’s data into 5-second chunks with the corresponding stress label to prepare the input for LaBraM. Because the dataset is highly imbalanced, we apply data augmentation to the elevated stress group to generate additional samples. We use a 75\% overlapping window (i.e., a step size of 250 samples, or 1.25 seconds) to create 5-second patches, increasing the elevated stress dataset size by roughly a factor of four. We do not augment the normal stress group.

Finally, to ensure the model effectively learns stress-related features rather than simply capturing temporal dependencies in EEG data, we shuffle the patches within each training, validation, and testing set. We export each patch as a pickle file, ready for input into LaBraM for fine-tuning. Table~\ref{label ratio} shows the statistics of the processed dataset.

\begin{table}[h]
    \centering
    \caption{Dataset Statistics}
    \label{label ratio}
    \begin{tabular}{lccc}
        \toprule
        \textbf{Set} & \textbf{Class} & \textbf{Ratio} & \textbf{Sample Count} \\
        \midrule
        \multirow{2}{*}{Train} & Increase & 0.5421 & 3594  \\
                               & Normal   & 0.4579 & 3036  \\
        \midrule
        \multirow{2}{*}{Validation} & Increase & 0.5594 & 480 \\
                                    & Normal   & 0.4406 & 378  \\
        \midrule
        \multirow{2}{*}{Test} & Increase & 0.4622 & 391  \\
                              & Normal   & 0.5378 & 455  \\
        \bottomrule
    \end{tabular}
\end{table}

% Train
% Label Ratios:
% Increase Class Ratio: 0.5421 (total: 6630, class count: 3594)
% Normal Class Ratio: 0.4579 (total: 6630, class count: 3036)

% Validation:
% Label Ratios:
% Increase Class Ratio: 0.5594 (total: 858, class count: 480)
% Normal Class Ratio: 0.4406 (total: 858, class count: 378)

% Test
% Label Ratios:
% Increase Class Ratio: 0.4622 (total: 846, class count: 391)
% Normal Class Ratio: 0.5378 (total: 846, class count: 455)

\subsection{LEM Fine-tuning}

We fine-tune LaBraM (\textasciitilde5.8M parameters) on our cleaned stress data patches. The full-parameter fine-tuning setup follows the TUAB dataset configuration described in the original LaBraM paper \cite{jiang2024large}, but uses a reduced learning rate and batch size. Table~\ref{tab:finetune} presents the detailed fine-tuning parameters. 

Note that the optimal learning rate and batch size can vary significantly for different datasets. For example, a learning rate of 5e-5 works well for the TUAB dataset, but for our stress dataset, it leads to overfitting within just 3 epochs. We run each training session for 50 epochs and save the checkpoint with the best classification accuracy. On average, 50 epochs take \textasciitilde2 hours on an NVIDIA RTX 2000 Ada GPU.

\begin{table}[h]
    \centering
    \caption{Fine-tuning Hyper-parameter Setup}
    \label{tab:finetune}
    \begin{tabular}{lc}
        \toprule
        \textbf{Hyper-parameter} & \textbf{Value} \\
        \midrule
        Weight Decay & 0.05 \\
        Batch Size & 32 \\
        Learning Rate & 1e-5 \\
        Epochs & 50 \\
        LR Warmup Epochs & 3 \\
        Layer Decay & 0.65 \\
        \bottomrule
    \end{tabular}
\end{table}

% \section{Results}
% We conducted three experiments to evaluate model robustness, the effectiveness of pre-training, and sensitivity to channel count. 

% The first experiment assessed the robustness of the fine-tuned model by testing it on different training, validation, and testing sets, each randomly generated using different seed values. Ideally, a leave-one-out training scheme would provide the most rigorous robustness evaluation; however, it is computationally prohibitive for full-parameter fine-tuning on Large EEG Models with our current compute resources. Instead, we randomly selected four seed values and recorded the results, as shown in Fig. \ref{fig:seeds}. Among these, Seed0 achieved the highest balanced accuracy of 85.39\% at epoch 35, followed by Seed42 with 83.68\% at epoch 34, Seed7 with 76.45\% at epoch 31, and Seed1 with 67.42\% at epoch 9. Some variation in accuracy is expected, as the small dataset size means the testing set typically consists of only two subjects from the elevated stress group, making it highly sensitive to out-of-distribution samples. Nevertheless, in all experiments, the accuracy significantly exceeded the chance level, demonstrating the model's robustness.

\section{Experiments and Results}

We conduct a sequence of experiments to evaluate model robustness, pre-training effectiveness, and model sensitivity to channel count.

\begin{figure}[htbp]
    \centering
    \includegraphics[width=0.48\textwidth]{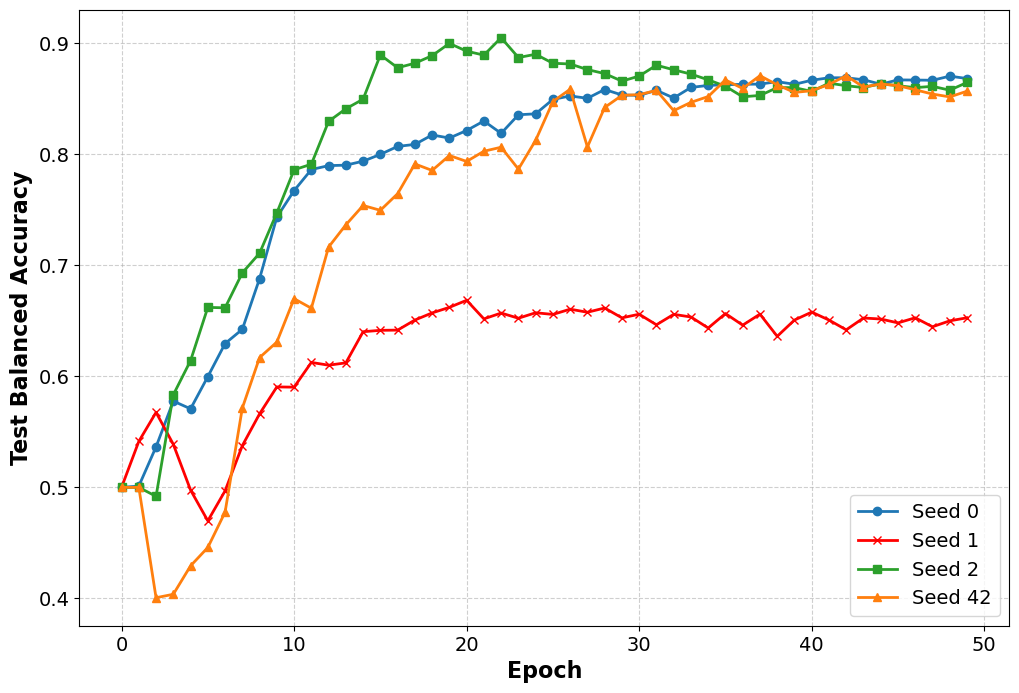}
    \caption{Finetuning results using 4 different data splitting seeds.}
    \label{fig:seeds}
\end{figure}

    The first experiment evaluates the robustness of the fine-tuned model by testing it on different training, validation, and testing sets, each generated using a different random seed. Ideally, a leave-one-out training scheme provides the most rigorous robustness evaluation; however, it is computationally prohibitive for full-parameter fine-tuning of LEMs with our current computing resources. Instead, we select four random seed values for data splitting and record the results, as shown in Fig.~\ref{fig:seeds}. The best models are trained based on the training set, selected from the validation set, and
    finally evaluated on the test set. The recorded best test balanced accuracy is selected from the model checkpoint that yields the best validation accuracy. Among these results, Seed 2 achieves the highest balanced accuracy of 90.47\% at epoch 23, followed by Seed 42 with 87.05\%, Seed 0 with 87.01\%, and Seed 1 with 66.84\%. We expect variation in precision, as the testing set typically has only two subjects in the elevated stress group because of the limited size of the dataset, making it highly sensitive to out-of-distribution samples. Nevertheless, in all experiments, the accuracy remains well above the chance level, demonstrating the model's robustness.

\begin{figure}[htbp]
    \centering
    \includegraphics[width=0.48\textwidth]{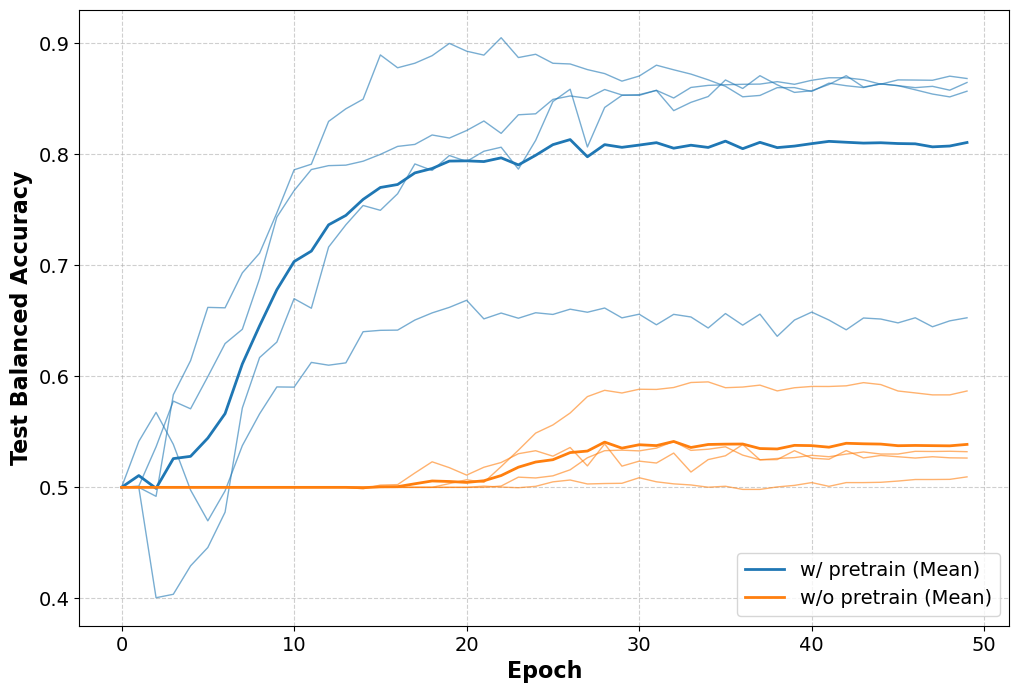}
    \caption{Fine-tuning results with pre-train v.s. without pre-train. Average results across different seeds are highlighted in darker colors}
    \label{fig:pre_nopre}
\end{figure}

We conduct the second experiment to validate the effectiveness of LEM pre-training. In this experiment, we fix the data splitting seed and compare the performance of a pre-trained model checkpoint with that of a newly initialized model. As shown in Fig.~\ref{fig:pre_nopre}, we observe a substantial performance gap in classification accuracy between the pre-trained and non-pre-training models. The averaged accuracy with pre-training reaches 81.04\%, whereas the model without pre-training only attains 53.76\%, which corresponds to nearly chance level. These results indicate that the model lacks sufficient prior information to learn meaningful patterns from a small dataset without pre-training.

\begin{figure}[htbp]
    \centering
    \includegraphics[width=0.48\textwidth]{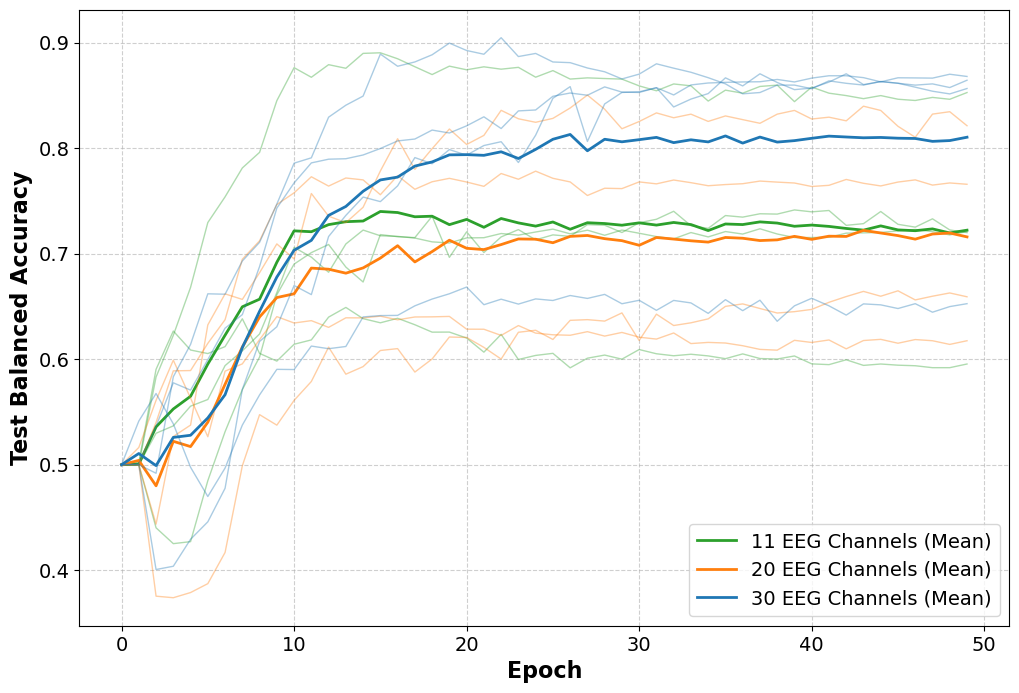}
    \caption{Fine-tuning results with different channel counts. Average results across different seeds are highlighted in darker colors}
    \label{fig:chanel_count}
\end{figure}
% Best test accuracy 30 channel(with channel interpolation): 83.68\%
% best accuracy 20 channel(no channel interpolation, only common channels are saved): 82.07\% 

We conduct a final experiment to test the sensitivity of LEM performance to the number of EEG channels. Minimizing the number of channels while maintaining reasonable performance is crucial for real-world applications. From the results, as shown in Fig.~\ref{fig:chanel_count}. We observe an averaged accuracy of 81.04\% occurs with 30 EEG channels. Reducing the channel count to 20 lowers the accuracy to 71.60\%, and using 11 channels yields 72.23\%. As expected, accuracy declines with fewer channels because of the reduced available information. The drop from 30 to 20 channels results in a decrease of over 9\% in accuracy. The comparable performance between 20 and 11 channels is not surprising, as some EEG channels carry more information about stress states than others, as demonstrated in \cite{chang2024online} and \cite{wang2022advancing}.

We select the 20-channel configuration purely at random \footnote{For readers' reference, the list of the 20 channels is: [FP1, F7, F3, F8, FZ, FC4, FT8, T3, C3, CZ, T4, P7, CP3, CPZ, CP4, T5, P3, PZ, P4, T6]}, whereas the 11-channel configuration follows \cite{chang2024online}, which carefully selects channels shown to yield the best results using a traditional machine learning classifier. Additionally, \cite{wang2022advancing} analyzes the power spectral density of all recording electrodes from the same stress dataset and finds that only specific frequency bands in a subset of channels exhibit statistical significance. Both studies suggest carefully selecting the most informative channels can enable comparable or even better results with fewer channels.

\section{Discussion}

Many researchers have attempted to develop stress classification models based on EEG data \cite{chang2024online}\cite{arsalan2019classification}\cite{hou2015eeg}. However, it is difficult to compare these studies because of different experimental designs and subject variability. A feasible direct comparison exists in \cite{chang2024online}, which uses the same dataset. As reported in \cite{chang2024online}, the highest balanced accuracy is 78.94\% using 11 channels and a 15-second window. For a direct comparison, we fine-tune LaBraM with the same set of channels and achieve a balanced accuracy of 72.23\% (with seed 42) using a 5-second window. Despite using a much shorter window length, our fine-tuned checkpoint outperforms \cite{chang2024online} by over 2\% when using 30 channels and underperforms by 6\% when using 11 channels. Moreover, the 5-second window configuration enables highly efficient inference and makes our checkpoint more suitable for near-real-time stress detection. 

Another notable aspect is that, compared to traditional machine learning models, fine-tuned LEMs require significantly more storage and computational resources, which may present challenges for deployment on wearable devices. In addition to computing resource requirements, LEMs—just like LLMs—operate as black boxes. Currently, very few studies \cite{chen2024eegformer} explore the interpretability of LEMs, leaving substantial room for further research. Nonetheless, our stress classification results validate the effectiveness of LEMs in capturing meaningful relationships in real-world EEG data. By sharing these preliminary findings, we aim to encourage more researchers in the BCI field to move beyond traditional task-specific expert-designed models and explore this new approach to BCI development. Furthermore, we anticipate continued growth and releases of LEMs from the BCI research community.

% Comparison with Chiyuan paper, advantage and disadvantage. 
% \begin{figure}[h]
%     \centering
%     \includegraphics[width=0.48\textwidth]{LDA_res.png}
%     \caption{11-channel classification result with respect to window length and step size from Chang et.al paper}
%     \label{fig:LDA}
% \end{figure}

\addtolength{\textheight}{-12cm}   % This command serves to balance the column lengths
                                  % on the last page of the document manually. It shortens
                                  % the textheight of the last page by a suitable amount.
                                  % This command does not take effect until the next page
                                  % so it should come on the page before the last. Make
                                  % sure that you do not shorten the textheight too much.

%%%%%%%%%%%%%%%%%%%%%%%%%%%%%%%%%%%%%%%%%%%%%%%%%%%%%%%%%%%%%%%%%%%%%%%%%%%%%%%%

%%%%%%%%%%%%%%%%%%%%%%%%%%%%%%%%%%%%%%%%%%%%%%%%%%%%%%%%%%%%%%%%%%%%%%%%%%%%%%%%

%%%%%%%%%%%%%%%%%%%%%%%%%%%%%%%%%%%%%%%%%%%%%%%%%%%%%%%%%%%%%%%%%%%%%%%%%%%%%%%%
% \section*{APPENDIX}

% Appendixes should appear before the acknowledgment.

% \section*{ACKNOWLEDGMENT}

% %%%%%%%%%%%%%%%%%%%%%%%%%%%%%%%%%%%%%%%%%%%%%%%%%%%%%%%%%%%%%%%%%%%%%%%%%%%%%%%%

% References are important to the reader; therefore, each citation must be complete and correct. If at all possible, references should be commonly available publications.

\bibliographystyle{unsrt}  % Ensures references appear in the order they are cited
\bibliography{ref}  % Matches the .bib file name (without the .bib extension)

% \bibitem{c1} Chang, Chi-Yuan, et al. "Online Mental Stress Detection Using Frontal-channel EEG Recordings in a Classroom Scenario." arXiv preprint arXiv:2405.11394 (2024).
% \bibitem{c2} Wang, Siwen. Advancing the neurophysiological understanding of stress, a study based on recorded Electroencephalography (EEG) data in real-world classroom. Diss. UC San Diego, 2022.
% \bibitem{c3} Arsalan, Aamir, et al. "Classification of perceived mental stress using a commercially available EEG headband." IEEE journal of biomedical and health informatics 23.6 (2019): 2257-2264.
% \bibitem{c4} Hou, Xiyuan, et al. "EEG based stress monitoring." 2015 IEEE international conference on systems, man, and cybernetics. IEEE, 2015.

\end{document}